\documentclass[conference]{IEEEtran}

\IEEEoverridecommandlockouts
\usepackage{cite}
\usepackage{amsmath,amssymb,amsfonts}
\usepackage{algorithmic}
\usepackage{graphicx}
\usepackage{textcomp}
\usepackage{xcolor}
\usepackage{soul}

\usepackage{dcolumn}
\usepackage{subfigure}
\usepackage{float}
\usepackage{caption}
\usepackage{tabularx}

\usepackage{stfloats}

\def\BibTeX{{\rm B\kern-.05em{\sc i\kern-.025em b}\kern-.08em
    T\kern-.1667em\lower.7ex\hbox{E}\kern-.125emX}}

\begin{document}

\title{Bandwidth-efficient distributed neural network architectures with application to body sensor networks\\}

\author{Thomas Strypsteen and Alexander Bertrand, \IEEEmembership{Senior Member, IEEE}

\thanks{This project has received funding from the European Research Council (ERC) under the European Union’s Horizon 2020 research and innovation
programme (grant agreement No 802895). The authors also acknowledge the financial support of the FWO (Research Foundation Flanders) for project G.0A49.18N, and the Flemish Government under the “Onderzoeksprogramma Artifici\"ele Intelligentie (AI) Vlaanderen” programme.
\newline
T. Strypsteen and A. Bertrand are with KU Leuven, Department of Electrical Engineering (ESAT), STADIUS Center for Dynamical Systems, Signal Processing and Data Analytics and with Leuven.AI - KU Leuven institute for AI, Kasteelpark Arenberg 10, B-3001 Leuven, Belgium (e-mail: thomas.strypsteen@kuleuven.be, alexander.bertrand@kuleuven.be).}
}

\maketitle

\begin{abstract}
In this paper, we describe a conceptual design methodology to design distributed neural network architectures that can perform efficient inference within sensor networks with communication bandwidth constraints. The different sensor channels are distributed across multiple sensor devices, which have to exchange data over bandwidth-limited communication channels to solve, e.g., a classification task. Our design methodology starts from a user-defined centralized neural network and transforms it into a distributed architecture in which the channels are distributed over different nodes. The distributed network consists of two parallel branches of which the outputs are fused at the fusion center. The first branch collects classification results from local, node-specific classifiers while the second branch compresses each node's signal and then reconstructs the multi-channel time series for classification at the fusion center. We further improve bandwidth gains by dynamically activating the compression path when the local classifications do not suffice. We validate this method on a motor execution task in an emulated EEG sensor network and analyze the resulting bandwidth-accuracy trade-offs. Our experiments show that the proposed framework enables up to a factor 20 in bandwidth reduction with minimal loss (up to 2\%) in classification accuracy compared to the centralized baseline on the demonstrated motor execution task. The proposed method offers a way to smoothly transform a centralized architecture to a distributed, bandwidth-efficient network amenable for low-power sensor networks. While the application focus of this paper is on wearable brain-computer interfaces, the proposed methodology can be applied in other sensor network-like applications as well.
\end{abstract}


\section{Introduction}
\label{section: Section1}

\subsection{Context and contributions}

In the last few years, technological advances such as miniaturization of microprocessors and energy-efficient batteries have increasingly enabled the usage of wearable, physiological sensors for ambulant health monitoring. Many applications however, will require recording of different data modalities or multiple channels of the same data type at different locations to extract meaningful patterns. This naturally leads to the concept of a body-sensor network (BSN), where the different sensors wirelessly share their data and solve a given task in a distributed fashion. Well-known applications include microphone arrays to detect heart and lung body sounds \cite{kirchner2017wearable}, electroencephalography (EEG) sensor networks \cite{bertrand2015distributed,seo2013neural} and other distributed or modular neuro-sensor platforms \cite{lee2021neural}. A major constraint in the design of these networks is that they should be energy-efficient, enabling a maximal battery lifetime. In BSNs, the typical energy bottleneck will be the wireless transmission of the data between the sensors and/or a fusion center \cite{reusens2009characterization, bertrand2015distributed,chen2012design}. Simply offloading all the recorded data to the cloud where they can be jointly processed will thus severely hamper the battery lifetime, presenting the need for different, bandwidth-efficient solutions \cite{bertrand2015distributed,somers2016removal}.
In this paper, we will present a framework to design deep neural network (DNN) architectures that deal with such bandwidth constraints. The framework is generic, in the sense that we make no prior assumptions on the DNN architecture itself. We start from a user-defined centralized neural network model for inference from multi-sensor input, and explain how this initial model can be used to build a distributed model that solves the same inference task. This conceptual methodology is then illustrated and analyzed in an EEG-based brain-computer interface task, which acts as a driver application driver throughout this paper.
\newline

EEG is a widely used, noninvasive way to measure the electrical activity of the brain. These signals can be harnessed for various purposes, including the monitoring and analysis of sleeping patterns \cite{de2017complexity}, epileptic seizure detection \cite{ansari2019neonatal},  the study of brain disorders after injuries \cite{giacino2014disorders} and brain-computer interfaces (BCI), which allows for direct communication between the human brain and external machines \cite{lawhern2018eegnet,allison2007brain,lotte2018review}. Traditional EEG requires patients to wear a bulky EEG cap with many wires that are connected to the acquisition device. This means that monitoring the patient's EEG can typically only be done in a hospital or laboratory environment. These limitations of classical EEG have led to a growing desire for ambulatory EEG, allowing for continuous neuromonitoring in daily life \cite{casson2010wearable}. A major enabler for these purposes is the development of mini-EEG devices: concealable, lightweight, miniaturized devices that are deployed behind or in the ear \cite{mirkovic2016target,mikkelsen2015eeg,bleichner2016identifying} or attached to the scalp \cite{baijot2021miniature,tang202034}. A single device would only be able to record one or a few EEG channels from its local area, hampering the performance in many of the previously mentioned applications. To mitigate this, multiple mini-EEG devices at different locations can be organized in a so-called wireless EEG sensor network (WESN) \cite{bertrand2015distributed,narayanan2019analysis,tang202034}. Each device can then perform some local processing on its own channels, before sharing its information with the other devices to perform the original, centralized EEG tasks in a distributed manner. This shift from one EEG cap towards a network of wireless, miniaturized devices affects the design of the machine learning models we use to perform these tasks in two major ways.
Firstly, to guarantee a comfortable user experience, we are only able to use a limited number of devices. We thus first need to solve an EEG channel selection task, determining \textit{how many} and \textit{where} these devices should be placed, minimizing the amount of devices, while maximizing the performance of the desired EEG task \cite{strypsteen2021end,narayanan2019analysis}. 
Secondly, the recorded channels are now stored on separate devices, meaning we cannot perform multi-channel processing without sharing the recorded data across the devices first. Simply transmitting the full, raw channels to a fusion center would incur enormous energy costs and severely hamper the battery life of the mini-EEG devices \cite{bertrand2015distributed,reusens2009characterization}. To achieve a viable battery life, we will thus need to limit the amount of data each device in the WESN can share and take this bandwidth constraint into account during the model design.
\newline

Recently, deep learning or DNN models have become more and more popular in the processing and analysis of physiological signals, including EEG \cite{roy2019deep}. This trend in combination with the shift towards low-power wearables cultivates a need to redesign such DNNs towards distributed architectures that can operate in modular sensor platforms, such as WESNs and other body-sensor networks. While a generic methodology for the first problem (i.e., channel selection and sensor placement for DNN-based inference) has been proposed in \cite{strypsteen2021end}, generic methodologies for the second problem (i.e. translating DNNs to bandwidth-efficient modular architectures) are still largely lacking. In this paper, we study how we can adapt existing \textit{centralized} DNN architectures to make them amenable for use in distributed settings with communication bandwidth constraints such as in body-sensor networks (and WESNs in particular). In the resulting distributed architecture, the sensor nodes learn to locally process the data, compress them to a desired degree and transmit them and finally fuse the compressed data to solve the desired task. To validate the applicability of this method, we study its performance on a motor execution EEG task and analyze the bandwidth-versus-performance tradeoff.
\newline

The main contributions of this paper are:
\begin{itemize}
    \item We introduce a design framework that maps a given centralized neural network architecture to a distributed architecture that is able to run efficiently on a bandwidth-constrained sensor network.
    \item We combine this framework with the early exit mechanism of \cite{teerapittayanon2017distributed} to further decrease the bandwidth by deciding on a per-sample basis how much data needs to be transmitted to the fusion center.
    \item We demonstrate the usage of our method by taking a centralized neural network architecture solving a given motor execution EEG classification task and decentralizing it. We analyze the resulting bandwidth-accuracy trade-offs of the resulting distributed architecture and demonstrate that with only small performance losses compared to the centralized baseline, substantial bandwidth gains can be achieved.
\end{itemize}

\subsection{Distributed deep learning: related work}

The literature of distributed deep learning is diverse and covers many different topics. A first class of methods distributes networks across multiple compute nodes, either to enable training of a single very large network that would otherwise not fit in memory on multiple standard CPU's or GPU's \cite{dean2012large} or to accelerate training by training a network on multiple devices in parallel and aggregating the gradient updates on each device \cite{goyal2017accurate}. A second line of research aims to map centralized models to a number of hardware devices to perform efficient inference. For instance, Bhardwaj et al. \cite{bhardwaj2019memory} employ multiple model compression techniques to map a single network to a number of smaller student networks with a limited memory footprint, while also minimzing the inter-device communication cost. Stahl et al. \cite{stahl2019fully} have a similar goal, but instead employs layer partitioning to perform the exact same operations as in the original network, but spread these out across devices.
\newline

All the previous work has one major factor in common, which makes them not applicable to our problem statement. They share the assumption that either all devices have access to all the input data or all the input data are generated in a central location and the energy of communicating this data to the worker nodes is not a constraint. The literature on deep learning where different channels or modalities of the input data itself is split across different devices is quite limited. The closest work to ours in this regard is the distributed deep neural network (DDNN) framework of Teerapittayanon et al. \cite{teerapittayanon2017distributed}. Similarly to our setting, the input data is distributed across devices. The local classifications of each device are aggregated and the confidence in this prediction is estimated with the normalized entropy of the resulting class probability distribution. If the confidence is high enough, this result - which only required the transmission of a classification vector of each node - is taken as the final result. Otherwise, a processed version of the data of each local device is forwarded to the cloud, thereby requiring a larger bandwidth. The main idea is thus to reduce bandwidth by only forwarding difficult samples that can't be correctly classified locally. Designing the optimal bandwidth-performance trade-off for a given application is then done by setting the desired confidence threshold of the local classification, with higher required confidence resulting in more data streamed to the cloud, but fewer misclassifications. In contrast, the main focus of this work will be to reduce bandwidth by designing an efficient architecture that compresses the data on our nodes to the desired degree, as will be described in the next section. However, both approaches are orthogonal to each other and we will ultimately combine them to gain even greater bandwidth gains without losing too much accuracy.

\subsection{Paper Outline}

The paper is organized as follows. In section \ref{section: Section 3} we introduce our framework and show how to combine it with the early exiting mechanism of \cite{teerapittayanon2017distributed}. Section \ref{section: Section4} presents the WESN use case, providing an overview of the used EEG dataset, how we emulate the environment of a WESN and design the neural network architecture for this specific use case. We then present our experimental results in section \ref{section: Section5} and finish with some conclusion in section \ref{section: Section6}.

\section{Proposed Method}
\label{section: Section 3}

In this section, we will conceptually describe the proposed bandwidth-efficient distributed architecture. We will first give a conceptual overview of the proposed architecture in Subsection \ref{section: Section3a}, while in Subsection \ref{section: architecture}, we will explain in more detail how a centralized neural network can be cast to this distributed architecture and how it is trained. In Section \ref{section: Section4} and \ref{section: Section5}, we will then apply this architecture design framework to a specific EEG inference task.

\subsection{Proposed architecture}
\label{section: Section3a}

\begin{figure}[!t]
    \centering
    \includegraphics[width = 0.5\textwidth]{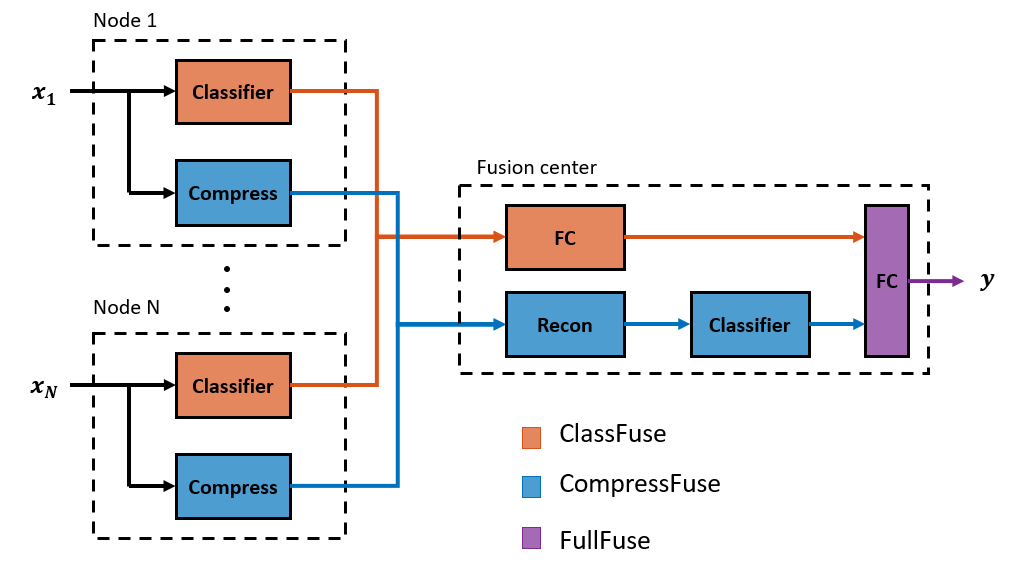}
    \caption{Illustration of the distributed neural network architecture and its modules. The orange \textit{ClassFuse} branch lets each node perform its own local classification with a single-channel neural network and combines these at the fusion center. The blue \textit{CompressFuse} branch compresses each channel locally and reconstructs the full multi-channel signal at the fusion center. This reconstruction is then classified by the multi-channel neural network. The purple \textit{FullFuse} module combines these two branches and performs the final classification.}
    \label{fig: architecturescheme}
\end{figure}

To build an architecture that minimizes the communication cost in a wireless sensor network, we propose a scheme where each local sensor (henceforth referred to as a node) compresses its recorded data as much as possible before sending it to a fusion center, where the final processing and inference will take place. The idea is to design an architecture that is able to interpolate between the two extreme cases of minimal and maximal communication, which also corresponds to minimal and maximal task accuracy. As the point of minimal communication, we take the setting where each node only transmits its local classification, as this would reasonably be the most condensed task-relevant information it could share. This setting, represented by the \textit{ClassFuse} branch (orange path in Figure \ref{fig: architecturescheme}), will serve as the basis of our architecture, with the other modules serving to trade extra bandwidth for an improved performance compared to this minimum-communication baseline. The point of maximal communication corresponds to each node simply transmitting its full recorded data. This would allow the fusion center to perform the same multi-channel processing as in the centralized case and achieve maximal accuracy. However, to achieve a trade-off between bandwidth and performance, we compress each signal at the local node, after which they are reconstructed at the fusion center. This is the task of the \textit{CompressFuse} branch (blue path in Figure \ref{fig: architecturescheme}). This \textit{CompressFuse} branch essentially mimics the original centralized network, although it operates on data that is distorted through the compression-reconstruction scheme. Finally, the results of the two branches are fused to provide a final output. A high-level schematic of such an architecture is illustrated in Figure \ref{fig: architecturescheme}. Another way to look at this architecture is as a combination of \textit{early} and \textit{late fusion}, respectively represented by the \textit{CompressFuse} and \textit{ClassFuse} branches. We will now delve deeper into the design of these modules for our WESN case and how they are trained.

\subsection{Design of the modules}

\label{section: architecture}

In this subsection, we will delve deeper into how we can use this framework to transform a given centralized architecture that has access to all input channels simultaneously, into a decentralized version that performs the same task.
\newline

\subsubsection{ClassFuse}

The task of the \textit{ClassFuse} branch (orange path in Figure \ref{fig: architecturescheme} is to let each node perform local classification and optimally fuse them together at the fusion center. The local classifications are performed with the original centralized architecture,where the input dimensions are reduced with respect to the number of local channels at each node. Each node then outputs the class scores as a log-probability vector to the fusion center. At the fusion center, these probability vectors are fused into a final class probability vector. This fusion is performed with a simple multilayer perceptron (MLP) with 1 hidden layer and Rectified Linear Unit (ReLU) nonlinearities on the concatenated outputs of the nodes. We take advantage of the modular nature of this network to train it in two stages. First, the weights of the local classifiers are pre-trained with a single-channel classification task. Then, the full \textit{ClassFuse} branch is trained end-to-end, with the weights of the local classifiers having a lower learning rate due to previously being pre-trained.  
\newline

\subsubsection{CompressFuse}

The task of the \textit{CompressFuse} branch (blue path in Figure \ref{fig: architecturescheme} is to compress each local recording, reconstruct the full multi-channel signal at the fusion center and classify this with the original, centralized neural network. To compress the local sensor channels at each node, we use two strided convolutional layers, with the value of the strides together determining the amount of compression (e.g. a stride of 2 and a stride of 3 resulting in a downsampling with a factor 6). We then upsample each channel separately with two transposed, strided convolutional layers, mirroring the strides of the compression layers. Another possibility would be to just drop the reconstruction alltogether, since this only introduces redundant information in the signal. However, when employing an existing neural network for classification, hyperparameters such as the length of the kernels have been tuned assuming a specific length of the time window at the input and a certain desired receptive field. Not employing reconstruction would break these assumptions and force us to redesign the original centralized classification network, which we aim to avoid as much as possible. Similary to the \textit{ClassFuse} branch, it is possible to perform the training of this branch in multiple stages. We could, for instance, first pre-train the compression-reconstruction layers as an auto-encoder by minimizing the mean squared error (MSE) between the reconstructed output and the input and then train the full \textit{CompressFuse} end-to-end. Whether this two-step training will be necessary, will largely depend on the size of the compression network compared to the classification network.
\newline

\subsubsection{FullFuse}
\label{section: fullfuse}
The \textit{FullFuse} module combines the classifications of the \textit{ClassFuse} and the \textit{CompressFuse} branches to perform the final classification. Similarly to the \textit{ClassFuse}, we simply use an MLP with 1 hidden layer and ReLU nonlinearity for this task. We train this module jointly with the previously trained \textit{ClassFuse} and \textit{CompressFuse}, once again employing a lower learning rate for the latter two. In Section \ref{section: Section4}, we will demonstrate that the output of \textit{FullFuse} obtains a higher accuracy than both the \textit{ClassFuse} and \textit{CompressFuse} branch separately.
\newline

In summary, the training of the network is thus comprised of the following steps:
\begin{enumerate}
    \item Train the single-channel local classifiers of each node.
    \item Train the full \textit{ClassFuse} branch, combining the local classifications and fine-tune the local classifier weights.
    \item Train the \textit{CompressFuse} branch, which compresses, reconstructs and classifies the node signals jointly.
    \item Train the entire network end-to-end, including the \textit{FullFuse} module, which combines the two previous branches to perform the final classification.
\end{enumerate}

In Subsection \ref{section: Section5_pretraining}, we will demonstrate the importance of using this piece-wise pre-training scheme, by comparing it to a direct end-to-end training from scratch.

\subsection{Early exiting}
\label{section: condex}

The \textit{ClassFuse} branch allows us to reach a certain, basis classifcation accuracy with minimal communication, while the \textit{CompressFuse} allows us to send additional information to boost this accuracy further. However, when the \textit{ClassFuse} is able to already correctly classify a substantial fraction of the samples on its own, this implies that for many of these samples, the extra information of the \textit{CompressFuse} branch is not necessary for a correct classification. Thus, we can save additional bandwidth by only transmitting the data for the \textit{CompressFuse} when we are not confident that the \textit{ClassFuse} has already successfully predicted the label of the current sample. This idea of allowing samples to exit the network early has previously been employed to reduce inference time \cite{teerapittayanon2016branchynet} and in the Distributed Deep Neural Network (DDNN) framework of \cite{teerapittayanon2017distributed} to decide whether a sample is processed locally or in the cloud. A common metric for classification confidence in this line of work, which we will employ here as well, is the normalized entropy of the softmaxed classification vector, defined as
\begin{equation}
    H(\boldsymbol{x}) = -\frac{1}{\log |C|} \sum_{i=1}^{|C|} x_{i} \log(x_{i}).
\end{equation}
with $|C|$ the number of classes and $\boldsymbol{x}$ a probability vector, which in this case is the softmaxed output vector of the \textit{ClassFuse} branch. Similar to \cite{teerapittayanon2017distributed}, we thus first perform classification using the \textit{ClassFuse} branch and measure the entropy of the current sample's output. If the entropy is lower than a certain threshold, we keep this output. If the entropy threshold is exceeded, we activate the \textit{CompressFuse} and combine this with the \textit{ClassFuse} to perform inference on the full network. The value of this threshold introduces a trade-off which can easily be tuned after network training: the higher we put this threshold, the less frequently the \textit{CompressFuse} branch will be activated, thereby saving more bandwidth at the cost of a reduced classification performance. A major advantage of combining early exiting with our bandwidth-efficient architecture design is that, in contrast to the compression factor of our \textit{CompressFuse} branch, the confidence threshold is a continuous parameter, thus allowing us to perform the bandwidth-accuracy trade-off in a continuous manner rather than a discrete one. The efficient architecture design on the other hand, allows us to start this trade-off from a more favorable point than we could otherwise.
\newline

\section{Case study: Wireless EEG Sensor networks}
\label{section: Section4}

In this section, we investigate the use of our distributed architecture in the context of a BCI task in a wireless EEG sensor network. We use data from a motor execution classification task, which is a well-known EEG-BCI paradigm for which large data sets as well as mature deep neural network architectures are available.
 
\subsection{Data set}
Motor execution is a widely used paradigm in the field of BCI. Real or intended body movement typically goes hand in hand with neuronal activity in certain motorsensory areas of the brain. The goal of motor execution is then to derive from these signals which movement was performed. In this work, we will employ the High Gamma Dataset \cite{schirrmeister2017deep}, containing about 1000 trials of executed movement following a visual cue, for each of the 14 subjects. The dataset also contains a separate test set of about 180 trials per subject, which we use to validate our results. The movements to be decoded are divided in 4 classes: left hand, right hand, feet and rest. While originally 128 channels were recorded for this dataset, we follow the approach of \cite{schirrmeister2017deep} and perform our experiments using only the 44 channels covering the motor cortex. The rest of our preprocessing procedure also follows the work of \cite{schirrmeister2017deep}:
\begin{itemize}
    \item Resampling to 250 Hz
    \item Highpass-filtering at 4 Hz
    \item Standardizing mean and variance per channel to 0 and 1
    \item Epoching in segments of 4.5 seconds, consisting of the 4 seconds after the visual cue and the 0.5 before.
\end{itemize}
The neural network architecture we employ for classification - and the one we will convert to a distributed architecture for our WESN - is the multiscale parallel filter bank convolutional neural network (MSFBCNN) proposed in \cite{wu2019parallel}. For completeness, a detailed summary of this network in table format can be found in Appendix A.
\newline

\subsection{WESN node emulation}
In traditional EEG caps, a channel is usually measured as the potential between an electrode at a given location and a common reference, typically the mastoid or Cz electrode. However, in the case of mini-EEG devices, we can only measure a local potential between two proximate electrodes belonging to the same device. To emulate this setting based on a standard cap-EEG recording, we follow the approach of \cite{narayanan2019analysis}. In this setting, each pair of electrodes within a preset maximum inter-electrode distance from each other is a candidate electrode pair or node we could measure. The signal this node records is then  emulated by subtracting one of the channels from the other, thus removing the common (far-distance) reference in the process. We applied this method with a distance threshold of 3 cm to our dataset, converting the 44 channels in 286 candidate electrode pairs or nodes. The resulting set of nodes had an average inter-electrode distance of 1.98 cm with a standard deviation of 0.59 cm.
\newline

\subsection{Node selection}

Since we are only able to use a limited number of mini-EEG devices, we will first perform a channel/node selection step to select the most relevant sensor nodes from the pool of 286 candidate nodes. To this end, we employ the regularized Gumbel-softmax method described in \cite{strypsteen2021end}. This method allows us to learn the $M$ optimal nodes for a given task and neural network by training said network jointly with a special selection layer that is able to learn the discrete variables involved in feature selection through simple backpropagation. The value of $M$ will also be varied throughout our experiments. We jointly train this selection layer of size $M$ with the centralized MSFBCNN architecture using the data from all subjects in the data set, resulting in a subject-independent set of $M$ mini-EEG nodes that are optimally placed to solve the motor execution task. The $M$ selected nodes are then used to design a distributed version of the MSFBCNN network as explained next.
\newline

\subsection{Distributed architecture design}

We build our distributed network by taking the MSFBCNN architecture as our centralized baseline. Thus, we employ a single-channel version of the MSFBCNN as our classifier on the local nodes and the multi-channel version as our classifier in the fusion center in the \textit{CompressFuse} branch (this corresponds to all the blocks denoted as 'classifier' in Figure \ref{fig: architecturescheme}). The MLP that fuses our local classifications in the \textit{ClassFuse} branch (the first orange FC block in Figure \ref{fig: architecturescheme}), consists of a simple MLP with one hidden layer of size 50 and ReLU nonlinearity in between. We use the same MLP architecture to fuse the output of the \textit{ClassFuse} and \textit{CompressFuse} (i.e. the purple FC block in Figure \ref{fig: architecturescheme}). The 'Compress' block consists of two convolutional layers, each consisting of a single kernel with strides to match a desired compression factor (e.g. one stride of 2 and one of 3 to achieve a compression factor 6). The 'Recon' block is built symmetrically to the 'Compress' block, with transposed convolutions replacing the normal convolutions. Since the reconstruction happens at the fusion center, it would also be possible to jointly reconstruct the channels using spatiotemporal filters instead, though our experiments indicated no advantage in this approach for our application.
\newline

The distributed network is trained using the data of all subjects jointly, using the procedure described in Section \ref{section: architecture}. Training is performed with the Adam optimzer \cite{kingma2014adam}, a learning rate of 0.001 and a batchsize of 64 for 50 epochs. Early stopping when the validation loss does not decrease for 5 epochs is employed to prevent overfitting. As soon as a layer has been trained for the first time, all subsequent fine-tuning of said layer will use a learning rate of only $10^{-4}$, a tenth of the original learning rate. When the full network has been trained, subject-dependent decoders are obtained by fine-tuning the full network end-to-end with subject-specific data.

\section{Experimental Results}
\label{section: Section5}

\subsection{Impact of short-distance nodes}

\begin{figure}[!ht]
    \centering
    \includegraphics[trim={0.5cm 0 0cm 0},width = 0.5\textwidth]{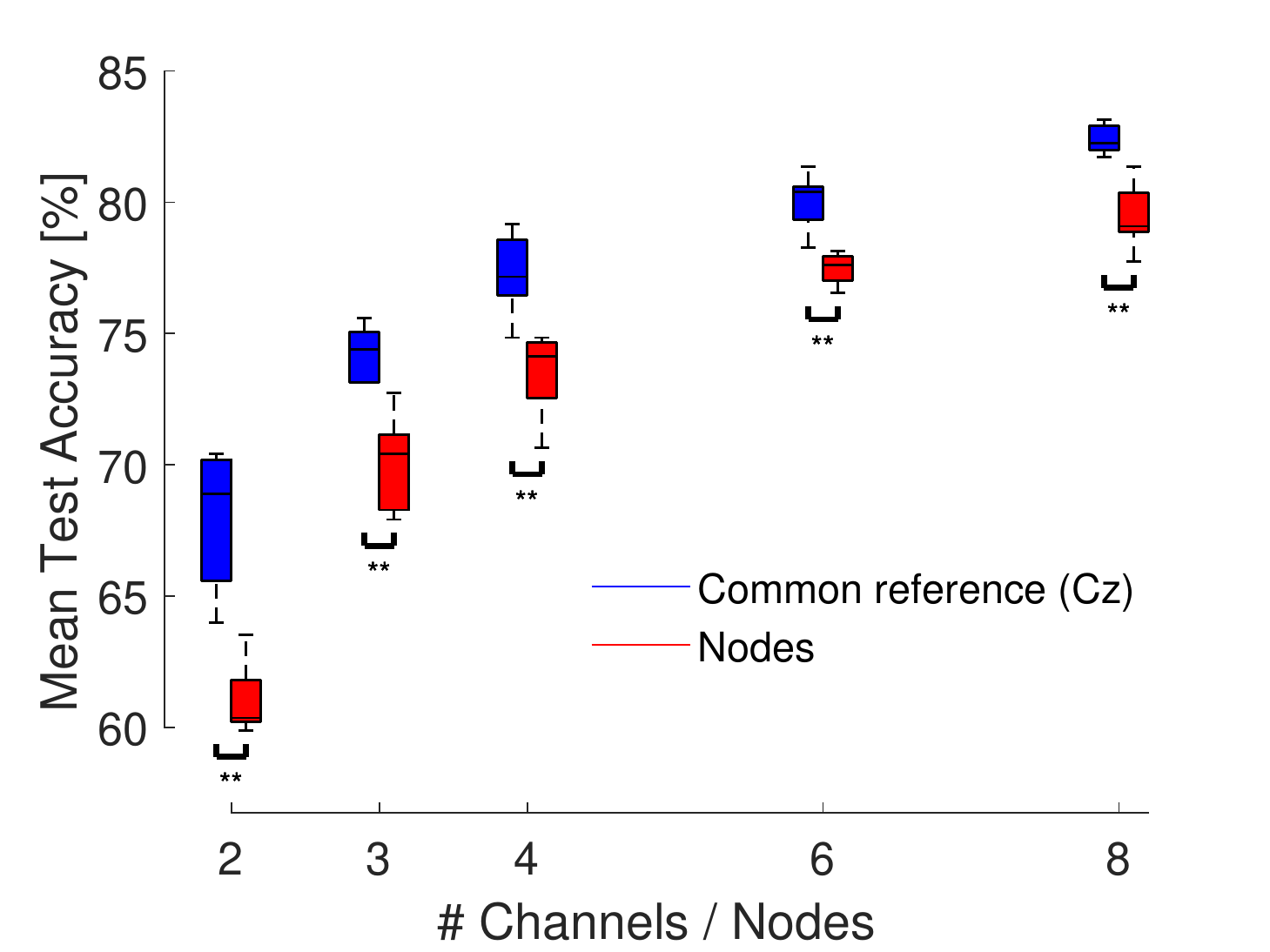}
    \caption{Comparison of the subject-dependent centralized motor execution accuracy when using $M$ short-distance nodes and $M$ Cz-referenced channels. Mean test accuracies across the subjects are plotted as a function the number of channels/nodes. The displayed boxplots are computed over 10 runs  and compared with independent samples t-test (no correction for multiple comparison). $**$ indicates statistically significant difference with $p <0.005$.} 
    \label{fig: channelsvsnodes}
\end{figure}

First, we take a look on how much using short-distance nodes instead of channels built from far-distance electrodes (with a common reference) impacts the accuracy of our motor execution task in the centralized case. Figure \ref{fig: channelsvsnodes} compares the subject-dependent accuracy of training the centralized baseline on the $M$ optimal mini-EEG nodes and the $M$ optimal Cz-referenced channels. Clearly, using electrodes only 2 to 3 centimeters apart from each other significantly affects the motor execution accuracy. These performance drops have also previously been observed in the field of auditory attention decoding \cite{narayanan2021eeg}, though only when the average distance between the electrodes becomes smaller than 3 cm. As observed in Figure \ref{fig: channelsvsnodes}, the difference between short-distance electrode pairs (nodes) and the original cap-EEG data tends to decrease when using more nodes, which is consistent with the observations in \cite{narayanan2021eeg,narayanan2019analysis}

\subsection{Distributed architecture}

\begin{figure*}[htbp]
    \centering
    \includegraphics[trim={1.6cm 0 -1cm 0},width=0.3\textwidth]{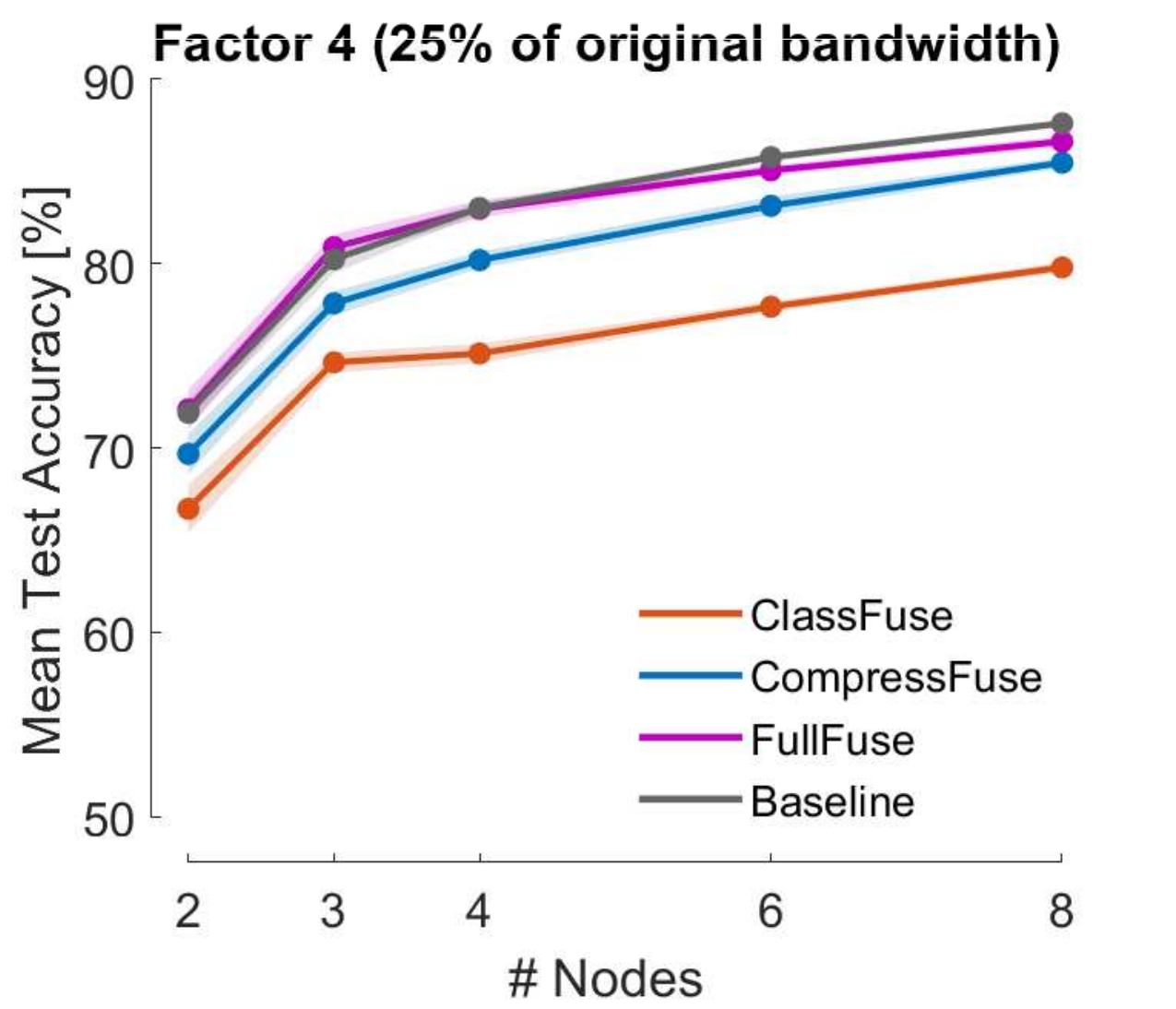}
    \includegraphics[trim={1.6cm 0 -1cm 0},width=0.3\textwidth]{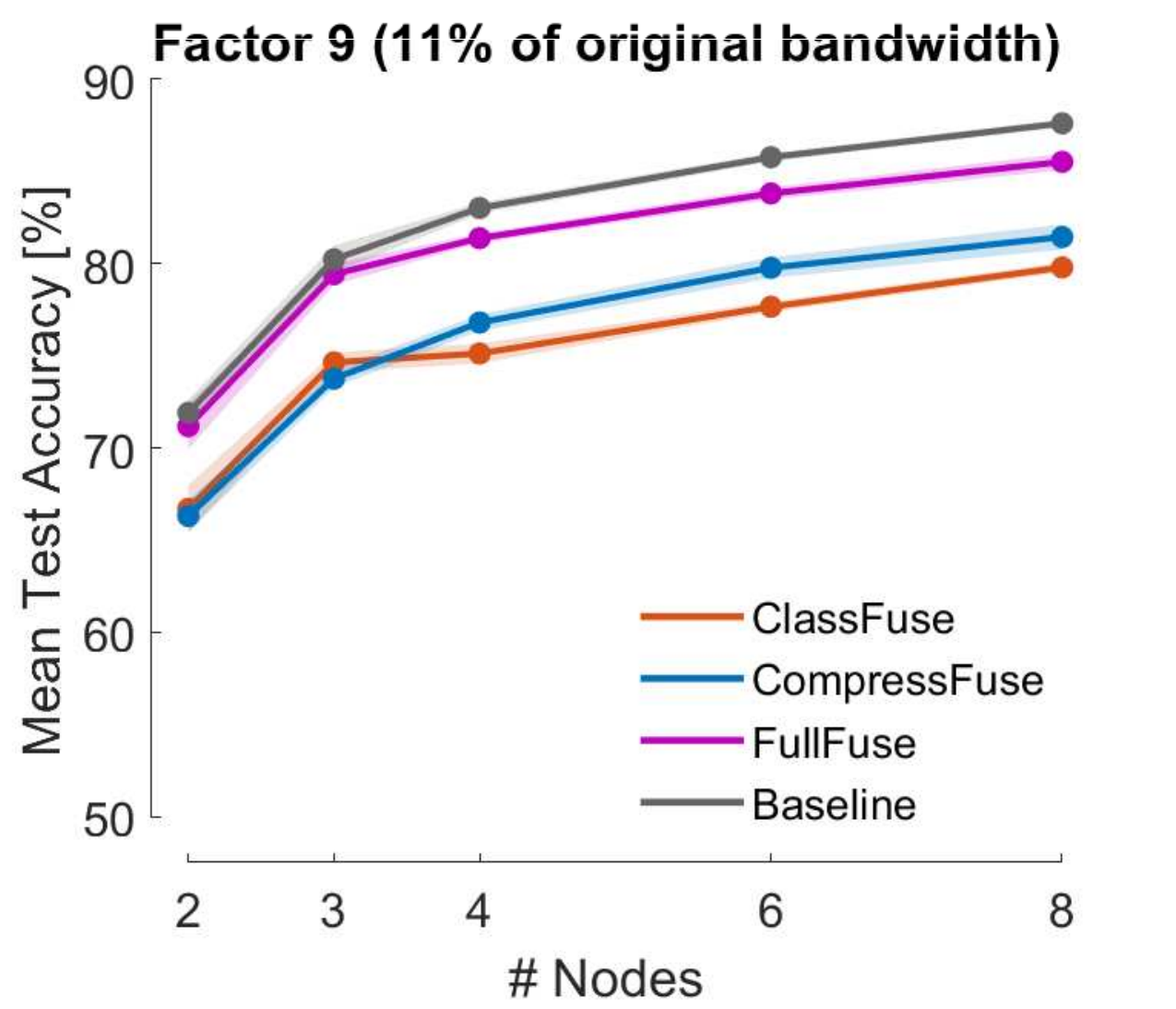}
    \includegraphics[trim={1.6cm 0 -1cm 0},width=0.3\textwidth]{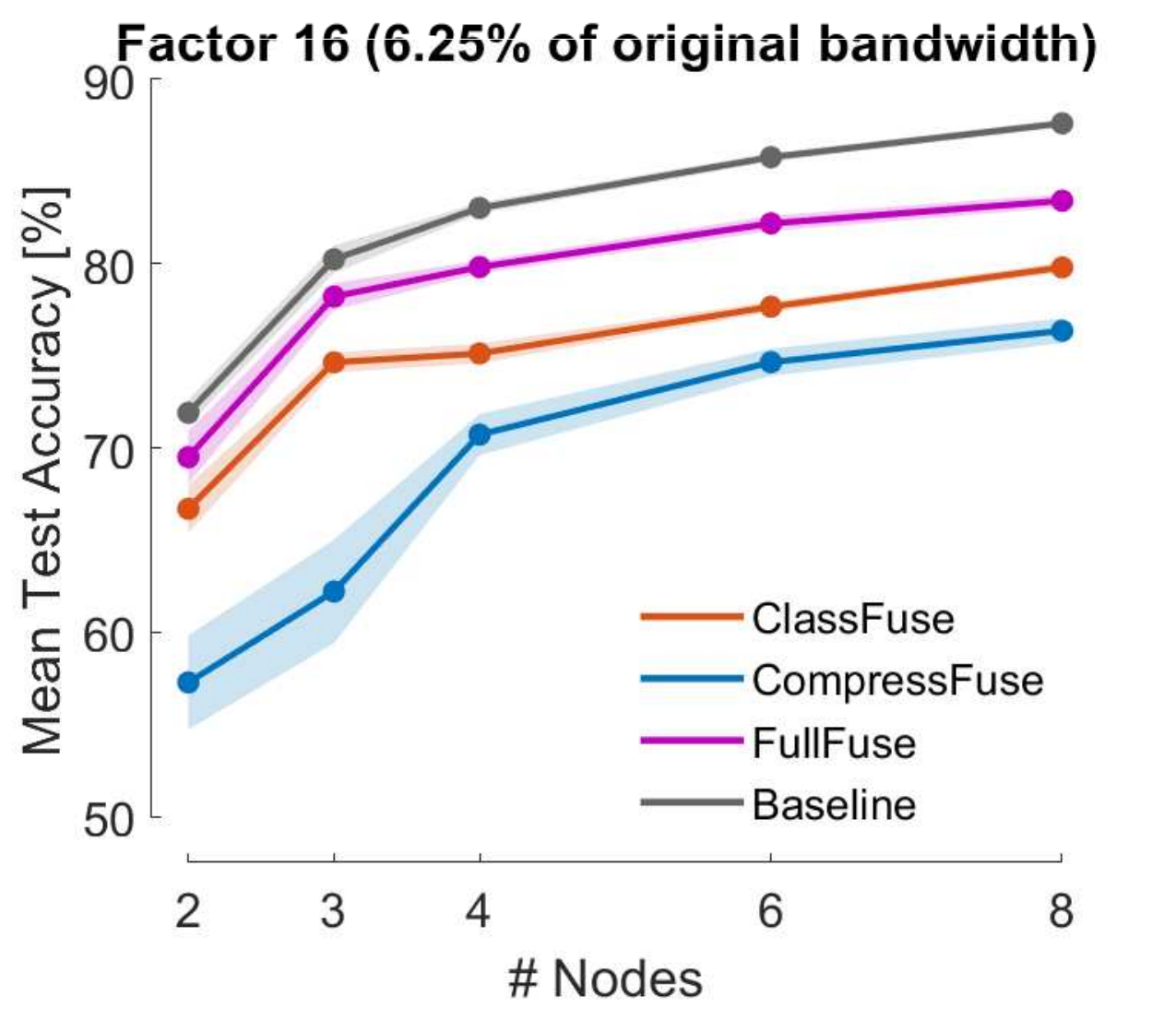}
    \caption{Comparison of the distributed architecture with different compression factors in the \textit{CompressFuse} branch. Each compression factor was obtained by two strided convolutions, with each stride equal to the square root of the compression factor. Mean test accuracies across the subjects are plotted as a function the number of nodes and averaged over 10 runs. Shades indicate standard error of the mean.}
    \label{fig: compressionperformance}
\end{figure*}

Next, we compare the performance of the proposed distributed architecture using different compression factors to the centralized baseline and investigate the individual and combined contribution of both branches. The results are summarized in Figure \ref{fig: compressionperformance}. A first observation is that, while the \textit{ClassFuse} is clearly less accurate than the centralized baseline, it still achieves reasonable accuracy considering it only requires the nodes to transmit a probability vector of size 4 (due to the 4-class task) compared to a full window of size 1125 (4.5 seconds sampled at 250Hz). A second observation is that the fusion of the \textit{ClassFuse} and \textit{CompressFuse} branches consistently and significantly outperforms the two separate branches, resulting in a \textit{FullFuse} that is competitive with the centralized baseline despite its much lower bandwidth usage. When moving to higher compression factors such as 16 however, the \textit{CompressFuse} has more and more trouble reconstructing the original EEG signal, especially at a lower number of nodes. At this point, its performance even drops below the \textit{ClassFuse} performance, while consuming more bandwidth. Remarkably, even at this stage it is still beneficial to fuse the two branches, suggesting that the information provided by the two branches is complementary. Thirdly, using more nodes results in a slowly increasing gap between the centralized baseline and the distributed architecture, since the unconstrained baseline is naturally more able to exploit the spatial correlations across the nodes. Finally, in terms of actual bandwidth gains, the efficient architecture design allows us to reach similar performance as the original network at 11\% of the original bandwidth and even with 6\% bandwidth, accuracy merely drops from 87\% to 82\% in the worst-case scenario.
\newline

\subsection{Impact of pre-training}
\label{section: Section5_pretraining}

\begin{figure}[!ht]
    \centering
    \includegraphics[trim={1.6cm 0 0cm 0},width = 0.5\textwidth]{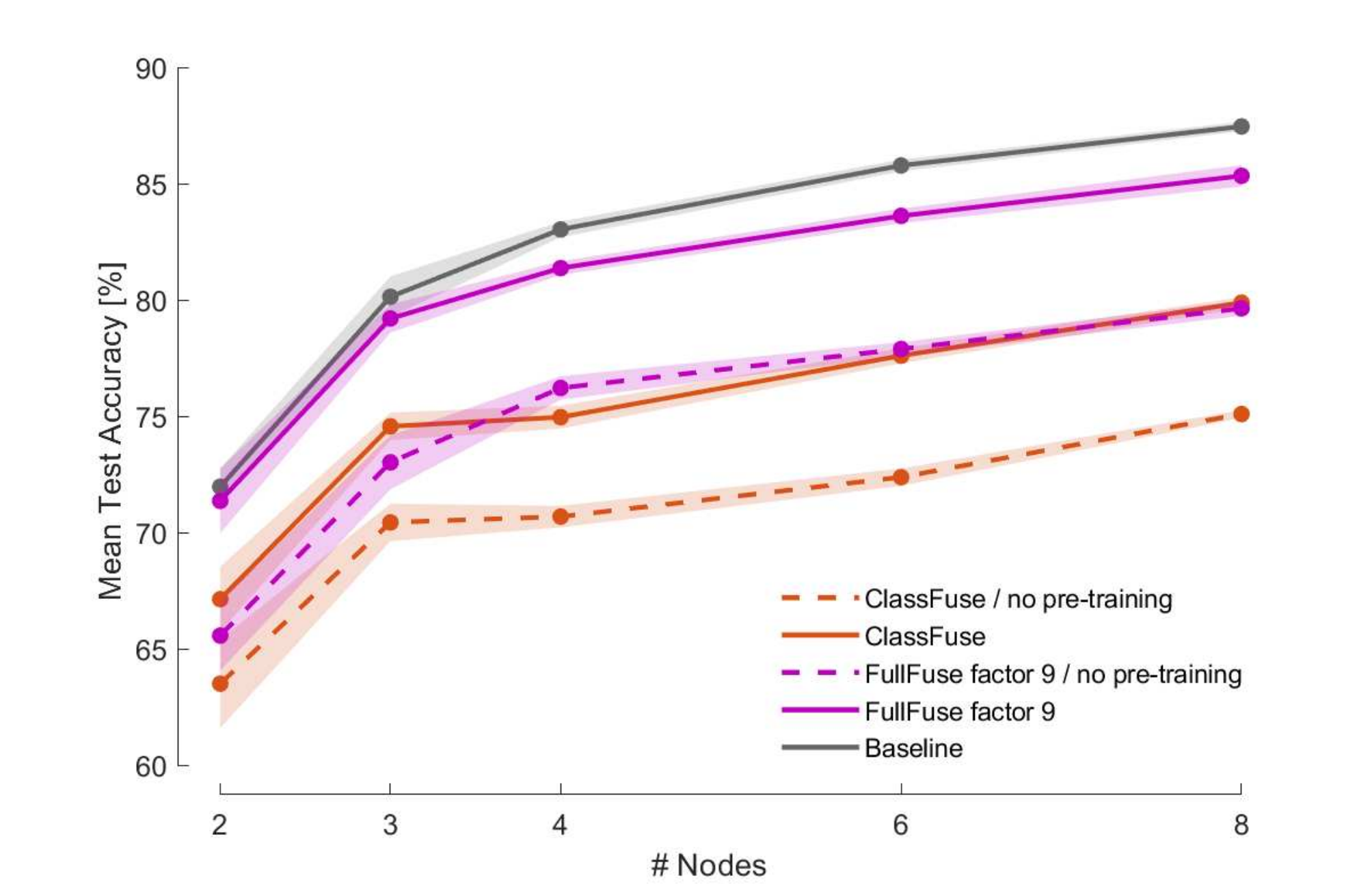}
    \caption{Effect of pre-training on the \textit{ClassFuse} and \textit{FullFuse} with compression factor 9 in the underlying \textit{CompressFuse} branch. Mean test accuracies across the subjects are plotted as a function the number of nodes and averaged over 10 runs. Shades indicate standard error of the mean.}
    \label{fig: pretrainingrelevance}
\end{figure}

To demonstrate the importance of the proposed training scheme, we also compare the performance of our network modules with and without this pre-training. As illustrated in Figure \ref{fig: pretrainingrelevance}, the network accuracy severely drops for the \textit{ClassFuse} and especially for the \textit{FullFuse} when training from scratch. This implies that the increased complexity of the distributed architecture indeed necessitates a custom training scheme taking advantage of its modular nature to train the network piece-wise. Though not shown Figure \ref{fig: pretrainingrelevance}, it should be noted that the \textit{CompressFuse} branch on the other hand, does not require pre-training at all, due to the small amount of parameters in the currently used compression-reconstruction layers. However, it stands to reason that pre-training in this branch might become necessary as well when deeper and more complex architectures are used in this branch.

\subsection{Early exiting}
\begin{figure*}[htbp]
    \centering
    \includegraphics[width=\textwidth,height=0.18\textheight,trim={1cm 0 0cm 0}]{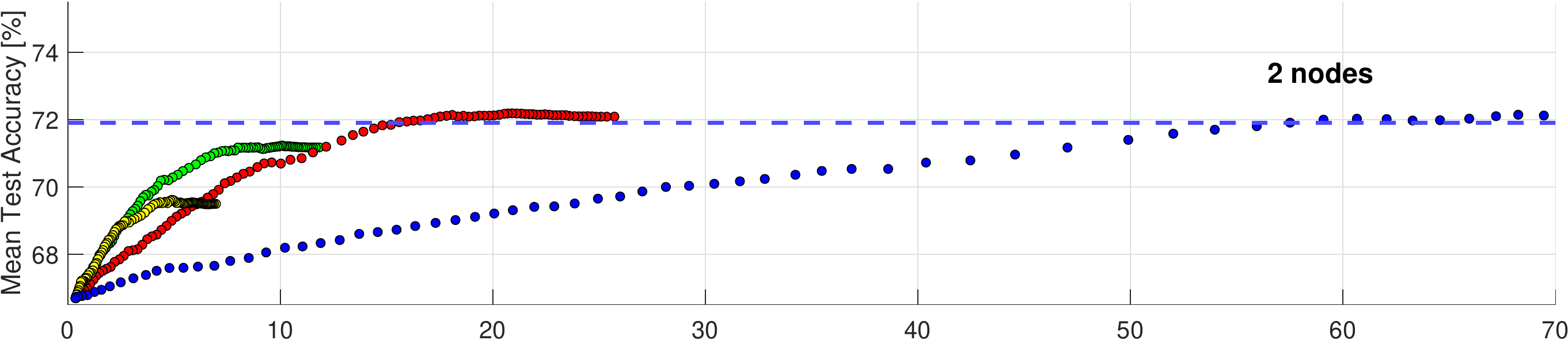}
    \includegraphics[width=\textwidth,height=0.18\textheight,trim={1cm 0 0cm 0}]{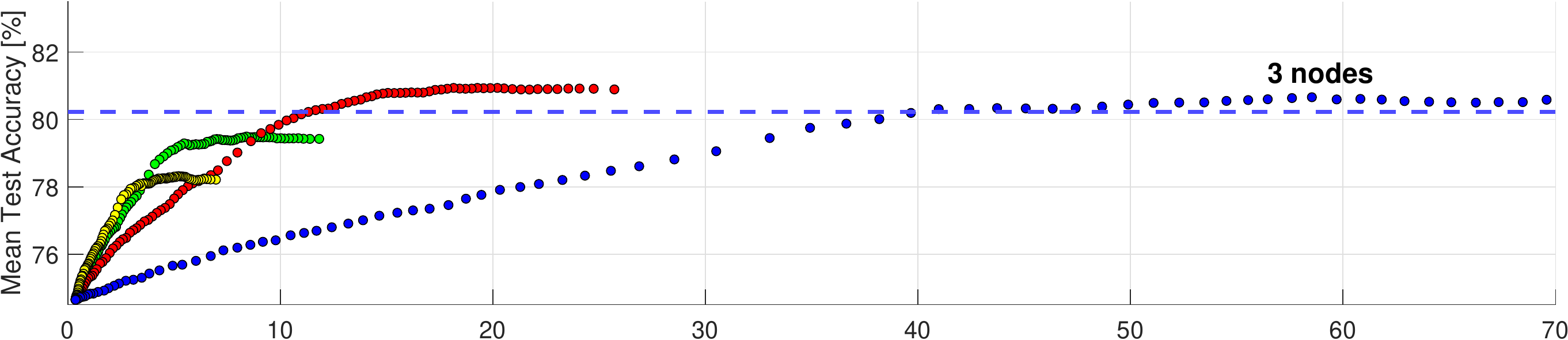}
    \includegraphics[width=\textwidth,height=0.18\textheight,trim={1cm 0 0cm 0}]{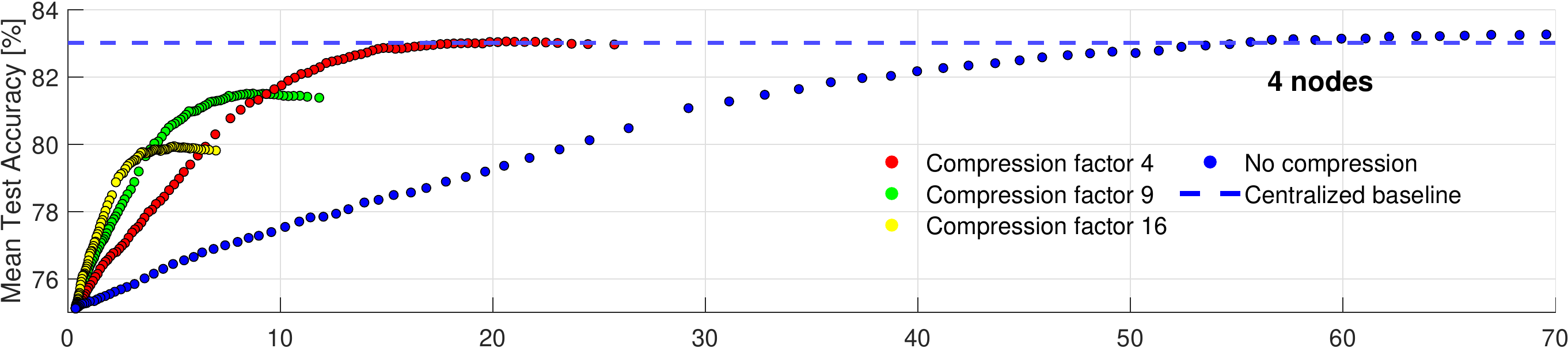}
    \includegraphics[width=\textwidth,height=0.18\textheight,trim={1cm 0 0cm 0}]{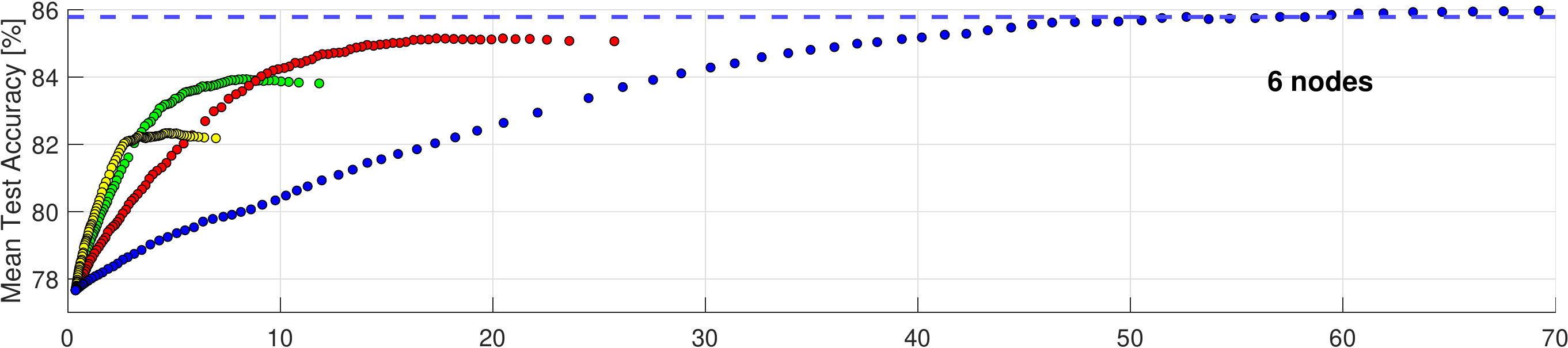}
    \includegraphics[width=\textwidth,height=0.18\textheight,trim={1cm 0 0cm 0}]{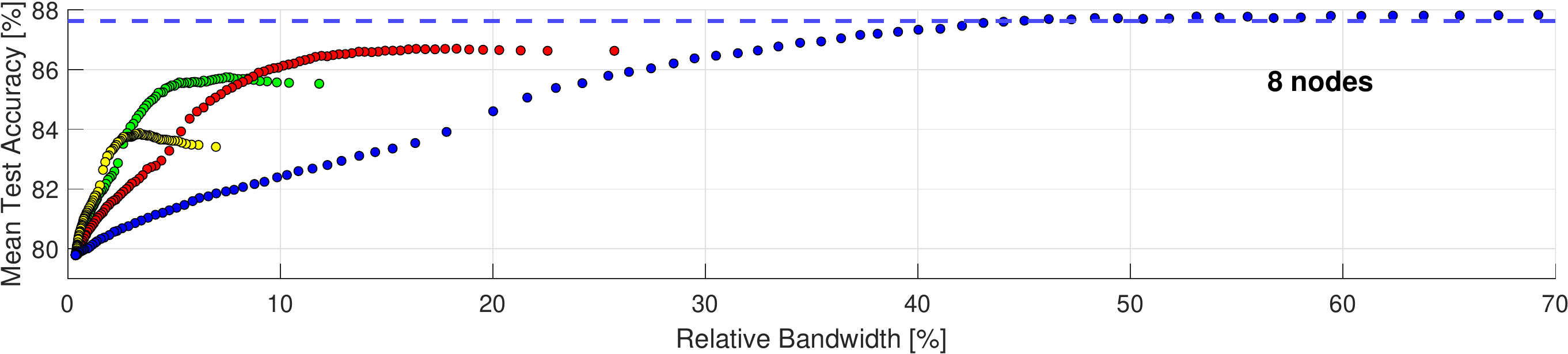}
    \caption{Bandwidth-accuracy trade-offs when applying early exiting to the distributed architecture with different compression rates of the \textit{CompressFuse} branch and for a different number of nodes. Bandwidth is measured as the average size of the data vector transmitted by each node relative to the full window size of each epoch, as determined by equation (\ref{eq: BW}). Mean test accuracies across the subjects are averaged over 10 runs. Dashed lines indicate centralized accuracy.}
    \label{fig: pareto}
\end{figure*}

Now that we have a more bandwidth-efficient architecture, we employ early exiting to let the network decide which samples are processed by the bandwidth-friendly \textit{ClassFuse} only and which by the complete \textit{FullFuse} network. By tuning the required confidence threshold between 0 (all samples are handled by the full network) and 1 (all samples are processed by \textit{ClassFuse} only) we can explore the accuracy-bandwidth trade-off in a continuous manner (instead of being confined to discrete non-prime compression factors) and find Pareto-optimal points, i.e., points where we cannot improve bandwidth or accuracy without sacrificing the other. We perform this trade-off for our distributed architecture with varying compression factors in Figure \ref{fig: pareto}. Each point in this plot corresponds to a network with $M$ nodes, compression factor $D$ and local exit confidence threshold $T$ (varied from 0 to 1 with a step size of 0.01), which in turn corresponds to a percentage of samples handled by the \textit{ClassFuse} alone, denoted by $\lambda(T)$. We compute the per-node bandwidth of this point, relative to the bandwidth required to run the centralized network (i.e. continuously transmitting the full recorded data window of length $L$ at each node). This relative per-node bandwidth B can be computed as:
\begin{equation}
\label{eq: BW}
    B(T) = \frac{1}{L} \left( |C| + (1-\lambda(T)) \frac{L}{D} \right).
\end{equation}
with $|C|$ the amount of classes, i.e. the size of the class probability vector (in this case 4).

A first observation to be made from the bandwidth-accuracy curves is that often, bandwidth can be reduced up to 50\% without any loss in accuracy. Interesting to note is that the deflection point at which the accuracy starts decreasing tends to shift more to the left, the more nodes we employ. This is not surprising, since more nodes implies a higher accuracy of the \textit{ClassFuse} branch, thus less samples for which the full network has to be activated. The advantage of using multiple nodes is thus twofold: it increases accuracy due to the higher amount of recorded data (see Figure \ref{fig: channelsvsnodes}), but also allows us to save more bandwidth \textit{per node} by requiring samples to pass through the whole network less often. Thus, instead of using nodes for increased accuracy, we can also employ them to save per-node bandwidth for the same accuracy. For instance, while using 3 nodes allows us to reach 80\% accuracy at 11\% of the original bandwidth, using 6 nodes allows us to do so at 1.3\% of the original bandwidth.
A second observation is that, when requiring low bandwidths, starting from a more bandwidth-efficient network with higher compression factors in the \textit{CompressFuse} branch and applying early exiting generally outperforms applying early exiting to a network with smaller compression factors. This is especially salient when comparing using \textit{CompressFuse} branches with compression (yellow,green and red in Figure \ref{fig: pareto}) to the version without compression (in blue), which corresponds to transmitting the raw sensor data and using the centralized baseline when the \textit{ClassFuse} is not confident enough. It is noted that these gains tend to decrease as we increase the compression factor,yet the non-compressive version remains outperformed by the compressive versions of \textit{CompressFuse}.
Finally, we can observe that the bandwidth gains of combining the efficient architecture design and the early exiting are substantial, allowing the network to operate at only 5\% of the original bandwidth, while never losing more than 2\% accuracy at that point. This demonstrates how the gains obtained by our proposed distributed architecture and those obtained by early exiting are complementary. 

\section{Conclusion and future outlook}

We have proposed a novel distributed neural network architecture design framework that can straightforwardly be mapped on a wireless sensor network and perform inference in this setting in a bandwidth-efficient manner. While we have applied it to the specific case of BCI in WESNs, the nature of this architecture is generic and can be applied to any kind of situation where the required input data is distributed across different sensor devices. This architecture consists of two parallel branches. The \textit{ClassFuse} branch lets each node in the network perform its own local classification and then aggregates these in a fusion center. The purpose of this late fusion procedure is to produce reasonable classifications while consuming the minimal amount of communication energy. To then be able to perform a trade-off between bandwidth and performance, the \textit{CompressFuse} branch compresses the recorded sensor signal of each node to a desired level and then approximately reconstructs the full multi-channel signal at the fusion center, where it can then be classified by a centralized network. These outputs of these two branches are then fused to perform the final classification.
To train this architecture, we have proposed a step-by-step procedure, taking advantage of the modular structure of the architecture to first pre-train every block separately. We have experimentally demonstrated both the need and the advantage of training the network in this way.
We have then combined the resulting network with the early exiting mechanism of \cite{teerapittayanon2017distributed} to decide on a per-sample basis whether to use the full network or the very bandwidth-friendly \textit{ClassFuse} to process the current input. We have shown that the introduction of the \textit{CompressFuse} branch allows to substantially push the Pareto-front upwards, in particular in low-bandwidth regimes.
\newline

We have validated the performance of our architecture on an emulated WESN solving a motor execution EEG task. We have used our architecture to obtain accuracy-bandwidth curves for this task, showing that for a realistic amount of nodes, we could save a factor 20 in bandwidth at the cost of 2\% mean test accuracy proving that good motor execution performances can be reached with both a low number of channels and a high reduction in the amount of data that needs to be transmitted from the nodes. An important observation in our experiments is the advantage of using multiple nodes in the sensor network. Not only does using more nodes increase classification accuracy, it also leads to a more favorable bandwidth-vs-accuracy trade-off, which in the case of WESNs implies an increased per-node battery life. In the future, we will explore ways to reach even higher reductions by using more sophisticated architectures for the \textit{CompressFuse} branch, which is currently a very simple model consisting of strided convolutions and transposed convolutions. We will also explore the generality of our findings on other EEG tasks, such as auditory attention decoding \cite{vandecappelle2021eeg} and epileptic seizure detection \cite{ansari2019neonatal} and other distributed platforms than WESNs.

\label{section: Section6}



\bibliographystyle{IEEEtran}
\bibliography{mybib.bib}

\onecolumn

\appendices

\section{MSFBCNN architecture}

\begin{table*}[!h]
\begin{tabular}{llllllll}
\hline
\textbf{Layer} & \textbf{\# Filters} & \textbf{Kernel} & \textbf{Stride} & \textbf{\# Params} & \textbf{Output} & \textbf{Activation} & \textbf{Padding} \\ \hline
Input          &                     &                 &                 &                    & (C,T)           &                     &                  \\
Reshape        &                     &                 &                 &                    & $(1,T,C)$       &                     &                  \\
Timeconv1      & $F_T$               & $(64,1)$        & $(1,1)$         & $64F_T$            & $(F_T,T,C)$     & Linear              & Same             \\
Timeconv2      & $F_T$               & $(40,1)$        & $(1,1)$         & $40F_T$            & $(F_T,T,C)$     & Linear              & Same             \\
Timeconv3      & $F_T$               & $(26,1)$        & $(1,1)$         & $26F_T$            & $(F_T,T,C)$     & Linear              & Same             \\
Timeconv4      & $F_T$               & $(16,1)$        & $(1,1)$         & $16F_T$            & $(F_T,T,C)$     & Linear              & Same             \\
Concatenate    &                     &                 &                 &                    & $(4F_T,T,C)$    &                     &                  \\
BatchNorm      &                     &                 &                 & $2F_T$             & $(4F_T,T,C)$    &                     &                  \\
Spatialconv    & $F_S$               & $(1,C)$         & $(1,1)$         & $4CF_TF_S$         & $(F_S,T,1)$     & Linear              & Valid            \\
BatchNorm      &                     &                 &                 & $2F_S$             & $(F_S,T,1)$     &                     &                  \\
Non-linear     &                     &                 &                 &                    & $(F_S,T,1)$     & Square              &                  \\
AveragePool    &                     & $(75,1)$        & $(15,1)$        &                    & $(F_S,T/15,1)$  &                     & Valid            \\
Non-linear     &                     &                 &                 &                    & $(F_S,T/15,1)$  & Log                 &                  \\
Dropout        &                     &                 &                 &                    & $(F_S,T/15,1)$  &                     &                  \\
Dense          & $N_C$               & $(T/15,1)$      & $(1,1)$         & $F_S(T/15)N_C$     & $N_C$           & Linear              & Valid            \\ \hline
\end{tabular}
\caption{Architecture of the MSFBCNN used for motor execution classification (the 'Classifier' blocks in Figure \ref{fig: architecturescheme}). In the model we use $T=1125, F_T=10, F_S=10$ and $N_C=4$. Each node operates a single-channel version version of this network where $C=1$ for the \textit{ClassFuse} and the fusion center contains a multi-channel version for the \textit{CompressFuse}, where $C$ is the number of nodes. This table is cited from \cite{wu2019parallel}.}
\end{table*}

\end{document}